\pdfoutput=1  

\documentclass[letterpaper, 10pt, conference]{ieeeconf}  

\IEEEoverridecommandlockouts                              

\usepackage{graphicx}
\usepackage{amsmath}
\usepackage{amssymb}
\usepackage{booktabs}
\usepackage{subcaption}
\usepackage{siunitx}
\usepackage{caption}
\usepackage{placeins}
\title{\LARGE \bf
Lightweight Generalized DeepFake Face Detection with WAVIE: Wavelet
Augmented Vision Intermediate Embeddings
}

\author{Arya Pulkit$^{*,1}$, Aditya Ruhela$^{*,1}$,  Akarshan Kapoor$^{*,1,2}$, Arnav Bhavsar$^{1}$\\%
\thanks{$^{*}$\hspace{2pt}Equal contribution from First Authors}%
\thanks{$^{1}$ School of Computing and Electrical Engineering, IIT Mandi, India}%
\thanks{$^{2}$ PHYTEC Messtechnik GmbH, Germany}%
\thanks{https://github.com/Kappuccino111/WAVIE-Master}%
\thanks{\copyright\ 2026 IEEE. Personal use of this material is permitted.
Permission from IEEE must be obtained for all other uses, in any current
or future media, including reprinting/republishing this material for
advertising or promotional purposes, creating new collective works,
for resale or redistribution to servers or lists, or reuse of any
copyrighted component of this work in other works.}%
}

\begin{document}

\maketitle
\thispagestyle{empty}
\pagestyle{empty}

\begin{abstract}

Deepfake detection systems often exhibit significant performance degradation when deployed on unseen manipulation methods, limiting their reliability in real-world multimedia environments. This lack of generalization poses critical challenges for misinformation mitigation, digital forensics, and human-centric AI systems. Existing detectors perform well on the forgery methods they are trained on, but their accuracy drops sharply on \emph{unseen} pipelines. To bridge this generalization gap, we propose \textbf{WAVIE} (\emph{W}avelet \emph{A}ugmented \emph{V}ision \emph{I}ntermediate \emph{E}mbeddings), an end-to-end architecture that combines complementary \textit{spatial} and \textit{frequency} cues on top of a frozen CLIP backbone \cite{radford2021learning}. WAVIE projects intermediate transformer embeddings through a lightweight learnable module, applies a \textbf{three-level Daubechies-6 (db6)} discrete wavelet transform (DWT), \emph{refines the low-frequency branch while preserving the high-frequency branch}, reconstructs the feature via inverse DWT, and performs classification. 

Trained \emph{only} on FaceForensics++~\cite{rossler2019faceforensics++}, WAVIE achieves \textbf{AUROC = 0.852} on Celeb-DF-v1~\cite{Li_2020_CVPR}, \textbf{0.852} on Celeb-DF-v2 and \textbf{0.831} on WildDeepFake (WDF) at the frame level, outperforming several state-of-the-art generalization baselines. Extensive ablation studies confirm the importance of both the wavelet module and the intermediate-feature aggregation for cross-dataset performance, highlighting the necessity of jointly leveraging spatial and frequency domains. These results position WAVIE as a strong baseline for deepfake detection in the wild.

\end{abstract}

\begin{figure*}[!t]
    \centering
    \includegraphics[width=1.0\linewidth]{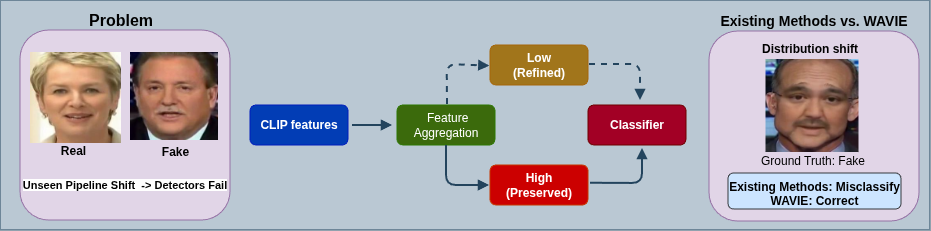}
    \captionsetup{justification=centering}
    \caption{Under unseen distribution shifts, the baseline model misclassifies samples, while WAVIE corrects predictions using low-frequency feature refinement.}
    \label{fig:wavie_front}
\end{figure*}

\section{INTRODUCTION}
\label{sec:intro}

Powerful generative models, from GANs to Diffusion frameworks\cite{rombach2022high}, now produce synthetic media that is hard to distinguish from camera imagery. While these models enable ease of creating media, they also power deepfakes that can depict people saying or doing things they never did, posing serious risks for misinformation, identity theft, and fraud. 

There are several deepfake detection proposals for preventing this, but the central obstacle in detecting deepfakes is \emph{domain shift}: detectors trained on one generator or dataset often fail on unseen pipelines because they latch onto the model or dataset-specific artifacts rather than universal cues of synthesis. As a result, in-domain accuracy can be deceptively high while cross-domain performance collapses. This reality raises an urgent need for \emph{generalizable} detectors that keep pace with rapidly evolving synthesis techniques.

Classical CNN detectors tend to overfit low-level patterns in the training distribution and may \emph{memorize} rather than learn transferable signals, sometimes even forming a de facto sink class for ''Real''~\cite{ojha2023towards}. By contrast, features from large, frozen vision transformers (e.g., CLIP ViT)\cite{radford2021learning} have shown strong transfer for synthetic-image detection with minimal training. However, most prior work relies primarily on the final embeddings, which prioritizes semantic content and can wash out subtle fabrication artifacts. This limitation suggests that more effective detection may require leveraging richer, multi-scale representations that preserve both fine-grained details and global context along with remaining robust across domains.

Motivated by this, we focus on two complementary aspects that are underexplored together: \emph{multi-layer representation aggregation} and \emph{frequency-aware feature processing}. The intermediate layers of a vision transformer capture a spectrum of information, ranging from local textures to global semantics, while frequency-based filtering can help in revealing structural inconsistencies that are less apparent in the spatial domain.



\textbf{WAVIE} builds on this perspective by integrating the components within a unified pipeline. We extract intermediate ViT features, learn per-layer, per-dimension importances to fuse them, and then apply a lightweight three-level wavelet split that refines the low-frequency statistics while preserving high-frequency detail. This hybrid spatial--frequency representation yields strong cross-dataset generalization in our experiments. Unlike adapter-based or per-layer projection designs, our implementation keeps a single shared projection per stage (before and after the wavelet block) and learns only a small fusion tensor and a low-band linear head, for a total of $\sim$2.14M trainable parameters. Thus, our work makes the following contributions:
\begin{itemize}
    \item We propose \textbf{WAVIE}, a novel detector that fuses intermediate ViT features via layer-wise feature aggregation and augments them with a representation-level low-band wavelet refinement.
    \item We demonstrate that WAVIE achieves strong cross-domain generalization, reaching 0.8518/0.8520 AUROC on Celeb-DF-v1/v2 and 0.8313 on WDF, competitive with recent state-of-the-art on Celeb-DF-v1 and best among listed methods on Celeb-DF-v2 and WDF, under an identical training pipeline and with only $\sim$2.14M trainable parameters atop a frozen backbone.
    \item We conduct extensive analyses, including ablations, and interpretability studies, highlighting the role of intermediate features, wavelet decomposition, and per-layer importance weights.
    \item We release code, models, and preprocessing scripts to foster reproducibility and further research in cross-domain deepfake detection.
\end{itemize}

\section{RELATED WORK}\label{sec:related_works}

\subsection{Deepfake Generation and Early Detection Strategies}
With the advancement in GAN and diffusion models, deepfakes have evolved from simple face-swapping into highly realistic synthetic media. Early works like FaceShifter~\cite{li2020advancing} highlighted the transition. It becomes increasingly difficult to distinguish synthetic images from the real ones with the improvement in the generation quality, not to mention the crucial need for performing well on unseen data.

\subsubsection{Early Deepfake Detection}
Traditional detection methods treated the problem as binary classification (Real vs. Fake) and relied on artifacts or "fingerprints" left by generative models~\cite{Yu_2019_ICCV,marra2019gans}. These methods, though being effective on known datasets, often struggled to generalize on unseen generators.

To improve robustness, researchers have explored patch-based classifiers, augmentation strategies, and frequency-domain analysis. Methods leveraging SRM filters and Fourier-based representations~\cite{frank2020leveraging,qian2020thinking,gong2025robust}, as well as F$^3$-Net\cite{qian2020thinking}, targeted high-frequency inconsistencies, while Face X-ray~\cite{li2020face} detected blending artifacts. Despite these advances, supervised detectors remained sensitive to dataset bias and showed poor cross-domain generalization~\cite{10.1007/978-3-030-58574-7_7}.

\subsection{CLIP and Vision Transformer (ViT) for Detection}
Recent approaches leverage large pre-trained models such as CLIP~\cite{radford2021learning}, which combines a text encoder with a ViT-based image encoder~\cite{dosovitskiy2020image}. Due to its training on diverse data, CLIP provides transferable representations suitable for deepfake detection.

Studies show that simple classifiers on frozen CLIP embeddings generalize well across generators~\cite{ojha2023towards}. However, relying solely on the final embedding emphasizes semantic information and may overlook subtle artifacts, motivating the use of intermediate ViT features.

\subsection{RINE: Leveraging Intermediate Transformer Features}
RINE~\cite{koutlis2024leveraging} utilizes intermediate ViT representations by extracting and concatenating the \texttt{[CLS]} token from multiple transformer layers. A Trainable Importance Estimator (TIE) assigns weights to each layer, combining low-, mid-, and high-level features.

With CLIP kept frozen and only lightweight modules trained, RINE achieves strong generalization across generators. However, it operates purely in the spatial domain and does not explicitly incorporate frequency information.

\subsection{Wavelet Based Deepfake Detection (Wavelet-CLIP)}
Frequency-domain methods capture artifacts that may not be visible in the spatial domain. Wavelet-CLIP~\cite{baru2025wavelet} applies a discrete wavelet transform (DWT) to the final CLIP embedding, separating low-frequency and high-frequency components.

The low-frequency component is refined using a small network, while the high-frequency component is preserved before reconstruction. Although effective, this approach relies only on the final embedding and does not leverage intermediate transformer features.

\section{METHODOLOGY: WAVIE}\label{sec:approach}

We propose \textbf{WAVIE}, a framework that performs representation-level spectral modeling over multi-layer transformer features for robust deepfake detection. Unlike methods that rely solely on final-layer embeddings, WAVIE leverages intermediate representations and their frequency characteristics within a unified pipeline.

A frozen CLIP ViT-L/14 backbone extracts the \texttt{[CLS]} token from each transformer block. WAVIE aggregates these features using a learnable layer-weighing mechanism in order to construct an improved feature vector.

We then apply a three-level Daubechies-6 (db6) discrete wavelet transform (DWT) on this fused representation. The decomposition separates low-frequency components (capturing global structure) from high-frequency components (capturing fine details). We refine only the low-frequency branch while preserving the high-frequency branch, and reconstruct the feature via inverse DWT (IDWT) before classification.

This design enables the model to balance stability and sensitivity: low-frequency refinement stabilizes global statistics, while preserved high-frequency components retain artifact cues. As a result, WAVIE improves generalization to unseen deepfake generators with minimal additional overhead.

\begin{figure*}[t]
  \centering
 \includegraphics[width=1.0\linewidth]{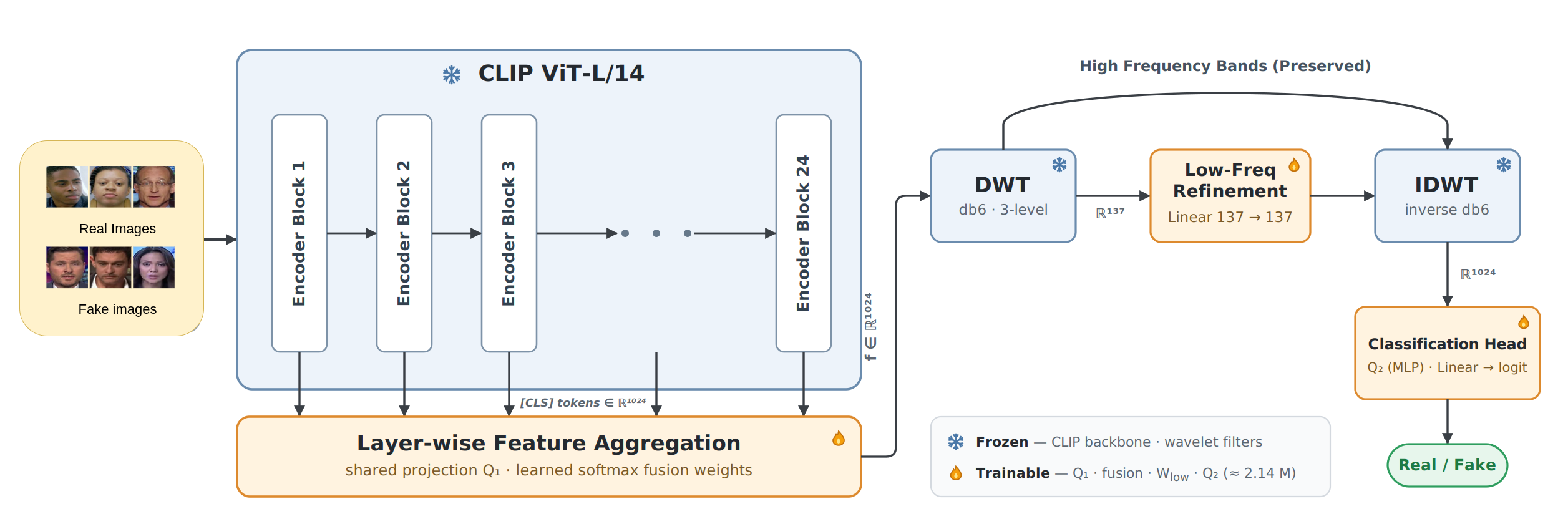}
\captionsetup{justification=centering,font=footnotesize}
\caption{WAVIE Framework: Multi-layer transformer features are aggregated, followed by representation-level wavelet decomposition, selective low-frequency refinement, and reconstruction for classification.}
\label{fig:wavie_full_model}
\end{figure*}

\textbf{Notation.} Let $n$ denote the number of transformer blocks in the CLIP-ViT backbone (for ViT-L/14, $n=24$). We index blocks by $l \in \{1,\dots,n\}$ and feature dimensions by $j \in \{1,\dots,d'\}$. Unless stated otherwise, we use a feature width of $d' = 1024$, corresponding to the hidden size of CLIP ViT-L/14 before the projection head.

1.\ \textbf{Feature Extraction (Frozen CLIP-ViT):} Given an input image $x \in \mathbb{R}^{3 \times 224 \times 224}$, we pass it through a pre-trained CLIP Vision Transformer, keeping all weights fixed. From each transformer block, we extract the \texttt{[CLS]} token representation. For ViT-L/14, each of these embeddings has dimension $1024$ (prior to the final projection to $768$, which we do not use). Collecting these across all $n$ blocks yields a matrix of intermediate features $G \in \mathbb{R}^{n \times 1024}$, capturing a range of information from low-level appearance cues in early layers to more abstract semantics in deeper layers.

2.\ \textbf{Per-Layer Projection ($Q_{1}$):} Each \texttt{[CLS]} feature $g_l \in \mathbb{R}^{1024}$ (for layer $l = 1, \dots, n$) is passed through a shared learnable projection $Q_{1}$ (lightweight MLP with Dropout, Linear, ReLU, Dropout), producing $h'_l \in \mathbb{R}^{1024}$. This step aligns features from different transformer layers, which otherwise vary in scale and distribution, into a common representation space. This improves stability during aggregation and helps combine complementary information across layers.

3.\ \textbf{Layer-wise Aggregation:} 
Taking the stacked projected features $H' \in \mathbb{R}^{n \times 1024}$ as input, we learn a set of scalar weights for each layer and feature dimension. Let $\alpha_{l,j}$ denote the learned logit for layer $l$ and feature index $j$. We normalize these across layers using a softmax:
\[
w_{l,j}=\frac{\exp(\alpha_{l,j})}{\sum_{m=1}^{n}\exp(\alpha_{m,j})}.
\]
The projected features are then aggregated as
\[
f_{j}= \sum_{l=1}^{n} w_{l,j}\,h'_{l,j}, \quad j=1,\dots,1024,
\]
resulting in a single fused feature vector $f \in \mathbb{R}^{1024}$. This allows the model to adaptively balance contributions from different layers, rather than relying on a fixed combination.

4.\ \textbf{Wavelet Decomposition:} 
Taking the fused feature $f \in \mathbb{R}^{1024}$ as input, we apply a 3-level Daubechies-6 (db6) discrete wavelet transform, separating it into low- and high-frequency components:
\[
f^{\text{low}}=W_{L}f, \qquad f^{\text{high}}_k=W_{H_k}f, \quad k=1,2,3.
\]
Here, $f^{\text{low}} \in \mathbb{R}^{137}$ captures coarse structure, while $f^{\text{high}}_1, f^{\text{high}}_2, f^{\text{high}}_3$ contain progressively finer details at each decomposition level. The wavelet filters are fixed, ensuring stable and invertible decomposition.

5.\ \textbf{Low-Frequency Refinement:} 
Taking $f^{\text{low}} \in \mathbb{R}^{137}$ as input, we update only the low-frequency component using a single linear layer:
\[
f^{\text{low}\prime}= W_{\text{low}}\,f^{\text{low}} + b_{\text{low}}, \quad W_{\text{low}} \in \mathbb{R}^{137 \times 137}.
\]
The high-frequency components $\{f^{\text{high}}_1, f^{\text{high}}_2, f^{\text{high}}_3\}$ are kept unchanged. This design is based on the observation that low-frequency signals tend to encode more stable structural patterns, while high-frequency components are more sensitive to noise and dataset-specific variations. Refining only the low-frequency part allows the model to focus on consistent cues while retaining fine details as auxiliary information.

6.\ \textbf{Reconstruction:} 
Taking the refined low-frequency component $f^{\text{low}\prime}$ and the original high-frequency components as input, reconstruction is performed using the inverse wavelet transform:
\[
\hat{f} = \text{IDWT}_{\text{db6}}\!\bigl(f^{\text{low}\prime}, \{f^{\text{high}}_{1}, f^{\text{high}}_{2}, f^{\text{high}}_{3}\}\bigr),
\]
yielding $\hat{f} \in \mathbb{R}^{1024}$. Since the db6 transform is invertible, the reconstruction preserves the overall structure of the representation.

7.\ \textbf{Final Projection and Classification:} 
The reconstructed feature $\hat{f} \in \mathbb{R}^{1024}$ is passed through a second learnable projection $Q_{2}$ (Dropout, Linear, ReLU, Dropout; output dimension $1024$), followed by a linear classification layer that maps $\mathbb{R}^{1024} \to \mathbb{R}^{1}$ to produce the final logit. Only the lightweight components $Q_{1}$, the low-frequency linear layer, $Q_{2}$, the classifier, and the layer-wise aggregation weights are trained, while the CLIP backbone and wavelet filters remain fixed. This keeps the number of learnable parameters relatively small (approximately $2.14$M for ViT-L/14 with $d' = 1024$).

8.\ \textbf{Training Objective:} 
The model is trained using a combination of binary cross-entropy and supervised contrastive loss:
\[
\mathcal{L} = \mathcal{L}_{\mathrm{BCE}} + \lambda\,\mathcal{L}_{\mathrm{SupCon}}.
\]
The binary cross-entropy loss is applied to the predicted probabilities $p_i$:
\[
\mathcal{L}_{\mathrm{BCE}}
= - \frac{1}{|\mathcal{B}|} \sum_{i \in \mathcal{B}}
\Big[ y_{i}\,\log(p_{i}) + (1-y_{i})\,\log(1-p_{i}) \Big],
\]
where $y_i \in \{0,1\}$ is the ground-truth label.

To encourage better separation between real and fake samples, we additionally apply a supervised contrastive loss on the normalized representation after $Q_{2}$. Let $z_i = \hat{f}_i / \|\hat{f}_i\|$ denote the $\ell_2$-normalized feature. For each sample $i$, positives are defined as other samples in the batch with the same label. The loss is given by:
\[
\mathcal{L}_{\mathrm{SupCon}} 
= \frac{1}{|\mathcal{B}|} \sum_{i \in \mathcal{B}}
\frac{-1}{|P(i)|} \sum_{p \in P(i)}
\log \frac{\exp\!\big( z_{i}^{\top} z_{p} / \tau \big)}
     {\sum\limits_{a \in \mathcal{B}\setminus \{i\}}
      \exp\!\big( z_{i}^{\top} z_{a} / \tau \big)} .
\]
We use a temperature $\tau = 0.07$ and set $\lambda = 0.1$ unless stated otherwise.

\textbf{Implementation Details:} We use CLIP's ViT-L/14 (24 blocks; per-block hidden width $1024$; the CLIP head projects to $768$ but is unused here). $Q_{1}$ and $Q_{2}$ are MLP mapping $1024{\to}1024$, each with ReLU and dropout $p{=}0.1$. A 3-level Daubechies-6 (db6) wavelet yields a low-frequency component of length 137 and corresponding high-frequency bands; the low-band is refined by a single linear layer mapping $137{\to}137$. We train with AdamW (lr $1{\times}10^{-3}$, weight decay $10^{-2}$), batch size $32$, and AMP; other training protocol details are in Sec.~\ref{sec:experimental-setup}.

\section{EXPERIMENTAL SETUP}\label{sec:experimental-setup}

\subsection{Datasets and Preprocessing}
We conduct experiments using FaceForensics++~\cite{rossler2019faceforensics++}, Celeb-DF-v1/v2~\cite{Li_2020_CVPR}, and WildDeepFake~\cite{zi2020wilddeepfake}. We use the raw (c0) variant of FaceForensics++ for training and validation, incorporating four manipulation types (\textit{Deepfakes}, \textit{Face2Face}, \textit{FaceSwap}, \textit{NeuralTextures}). To mitigate class imbalance, we sample real videos at $1$ fps and fake videos at $4$ fps. Each extracted frame is processed by MTCNN (\texttt{facenet\_pytorch} v2.5): we select the \emph{largest detected face} per frame, expand the bounding box by 20 pixels (when in-bounds), crop, and resize to $224{\times}224$ pixels. Frames with no detected face are discarded. This yields $131{,}155$ real and $175{,}465$ fake frames, randomly split into training and validation subsets (80\%/20\%) with a fixed seed. Detailed counts appear in Table~\ref{tab:datasets}. We deliberately focus on FF++$\rightarrow$Celeb-DF/WDF transfer to isolate the effects of intermediate-layer fusion and wavelet refinement, avoiding confounds from multi-dataset training. All splits use the same pipeline: after face cropping and resizing, images undergo CLIP's standard RGB normalization; no additional augmentations (e.g., color jitter, blur) are applied, in order to isolate the effect of the representation and wavelet stages.

\begin{table}[!htb]
\centering
\small
\begin{tabular}{llrrl}
\toprule
Dataset & Split & Real & Fake & Eval. \\
\midrule
FF++        & Train & 104,924 & 140,372 & -- \\
            & Val   & 26,231  & 35,093  & Self \\[0.3em]
CDF-v1      & Test  & 8,073   & 16,302  & Cross \\
CDF-v2      & Test  & 10,165  & 21,458  & Cross \\
WDF      & Test  & 58,659  & 107,003 & Cross \\
\bottomrule
\end{tabular}
\captionsetup{font=footnotesize}
\caption{Dataset statistics and evaluation role.}
\label{tab:datasets}
\end{table}

\subsection{Network Architecture and Hyperparameters}
Complete architectural details are in Section~\ref{sec:approach}. Briefly, the detector employs a \emph{frozen} CLIP ViT-L/14 backbone. We extract per-block \texttt{[CLS]} tokens (pre projection, width $1024$), project them to $1024$, fuse them via a Trainable Importance Estimator, and apply a 3-level Daubechies-6 (db6) wavelet with a single linear low-band refinement ($137{\to}137$) before classification. The total number of trainable parameters is approximately $2$M.
\begin{table*}[!t]
\centering
\begin{tabular}{l l l S S S S }
\toprule
{Model} & {Venue} & {Backbone} &
{   FF++} & {Celeb-DF-v1} & {Celeb-DF-v2}& {WDF}  \\
\midrule
EfficientNet\cite{tan2019efficientnet}  & ICML'19   & EfficientNet-B4 & 0.990 & 0.790 & 0.748 & \\
Xception\cite{chollet2017xception}      & ICCV'19   & Xception        & 0.990 & 0.779 & 0.736 & 0.652 \\
F$^3$-Net\cite{qian2020thinking}        & ECCV'20   & Xception        & 0.990 & 0.776 & 0.735 & 0.677 \\
Face~X-ray\cite{li2020face}             & CVPR'20   & HRNet           & 0.970 & 0.709 & 0.678 &       \\
CLIP \cite{ojha2023towards}             & CVPR'23   & ViT-L/14        & 0.960 & 0.743 & 0.750 &       \\
LAA\cite{peng2024local}                 & Neural Networks'24 & ConvNeXt V2-B   & 0.993 &       & 0.779 & 0.709 \\
IAL\cite{zhou2025identity}              & ICASSP'25 & EfficientNet-b4 & 0.993 & 0.796 & 0.771 &       \\
Wavelet-CLIP\cite{baru2025wavelet}          & WACV'25   & ViT             &       & 0.756 & 0.759 &       \\
WaViT-CDC\cite{badr2025wavit}           & OJSP'25   & ViT             & 0.988 &       & 0.816 & 0.812 \\
RSG-DA\cite{gao2025learning}            & TDSC'25   & EfficientNet-B4 & 0.971 & \bfseries 0.860 & 0.799 &       \\
CDA-Net\cite{li2025mamba}               & SMC'25    & EfficientNet    & 0.981 &       & 0.820 & 0.774 \\

\textbf{WAVIE (ours)} & -- & Frozen ViT-L/14 & 0.980 &  0.852 & \bfseries 0.852 & \bfseries 0.831  \\

\bottomrule
\end{tabular}
\captionsetup{font=footnotesize}

\caption{Frame-level AUROC on self-domain (FF++) and cross-domain (Celeb-DF, WDF) benchmarks.}
\label{tab:perf}

\end{table*}

\subsection{Training Protocol}
We train for 20 epochs using AdamW with learning rate $1{\times}10^{-3}$ and weight decay $10^{-2}$, batch size $32$, and automatic mixed precision (AMP). A OneCycleLR schedule warms up and then decays the learning rate across epochs. We also maintain an exponential moving average (EMA) of model weights with decay $0.999$. The loss is
\begin{equation*}
\mathcal{L} = \mathcal{L}_{\mathrm{BCE}} + 0.1\,\mathcal{L}_{\mathrm{SupCon}},
\end{equation*}
where the supervised contrastive term is computed from a \emph{single} view per frame using same-class positives within the batch (temperature $\tau{=}0.07$). We do not use class re-weighting, since the sampling strategy (1 fps real / 4 fps fake) already controls imbalance ($131{,}155$ vs.\ $175{,}465$). We fix random seeds (Python/NumPy/PyTorch) and enable cuDNN deterministic operations for reproducibility.

\subsection{Evaluation Protocol}
We evaluate WildDeepFake\cite{zi2020wilddeepfake}, Celeb-DF-v1 and v2~\cite{Li_2020_CVPR} separately, maintaining the same preprocessing as in training. The primary metric is \emph{frame-level} AUROC. Unless stated otherwise, all reported numbers are frame-level. We focus on cross-dataset tests (trained on FF++, tested on Celeb-DF and WildDeepFake) as the most stringent evaluation.


\section{RESULTS}

\subsection{Comparison with State-of-the-Art}
This subsection compares the proposed method with existing approaches under cross-domain evaluation settings. Table~\ref{tab:perf} presents results on the \textit{FaceForensics++} (FF++) self-domain benchmark and the cross-domain \textit{Celeb-DF-v1}, \textit{Celeb-DF-v2}, and \textit{WDF} datasets. All results are reported in terms of frame-level AUROC.

\paragraph{Self vs.\ Cross Domain.}
Most methods achieve near-saturated performance on FF++, with several models exceeding 0.99 AUROC (e.g., \emph{EfficientNet}, \emph{Xception}, \emph{F$^3$-Net}, \emph{IAL}). However, their performance drops substantially on unseen datasets. For instance, \emph{Xception} achieves 0.779/0.736 on Celeb-DF-v1/v2, and \emph{Face X-ray} further degrades to 0.709/0.678, highlighting the difficulty of cross-domain generalization.

\paragraph{Recent Strong Baselines.}
Recent approaches improve robustness but still exhibit performance gaps. \emph{RSG-DA} achieves the best result on Celeb-DF-v1 (0.860), while \emph{CDA-Net} (0.820) and \emph{WaViT-CDC} (0.816) perform strongly on Celeb-DF-v2. On WDF, \emph{WaViT-CDC} reaches 0.812, representing one of the strongest baselines on this dataset.

\paragraph{WAVIE.}
The proposed \textbf{WAVIE} achieves 0.980 AUROC on FF++, 0.8518 on Celeb-DF-v1, 0.8520 on Celeb-DF-v2, and 0.8313 on WDF. Compared with recent state-of-the-art methods, WAVIE is competitive with the best-performing model on Celeb-DF-v1, while achieving the \textbf{highest performance on Celeb-DF-v2 and WDF} among all listed methods.

In particular, WAVIE improves over strong recent baselines such as \emph{CDA-Net} (+3.2 on v2, +5.7 on WDF) and \emph{WaViT-CDC} (+3.6 on v2, +1.9 on WDF), demonstrating enhanced robustness to unseen manipulation pipelines.

Notably, WAVIE also substantially outperforms the frozen \emph{CLIP} baseline (+10.9/+10.2 on v1/v2) and \emph{Wavelet-CLIP} (+9.6/+9.3), confirming that the layer-wise feature aggregation and frequency refinement are complementary.

Practically, WAVIE achieves consistently strong cross-domain performance while keeping the backbone frozen and the trainable budget low ($\sim$2M), suggesting that representation-level wavelet refinement combined with multi-layer fusion provides an effective and parameter-efficient approach to generalizable deepfake detection.

\begin{figure}[ht]
\centering
\begin{subfigure}[b]{0.45\linewidth}
    \includegraphics[width=\linewidth]{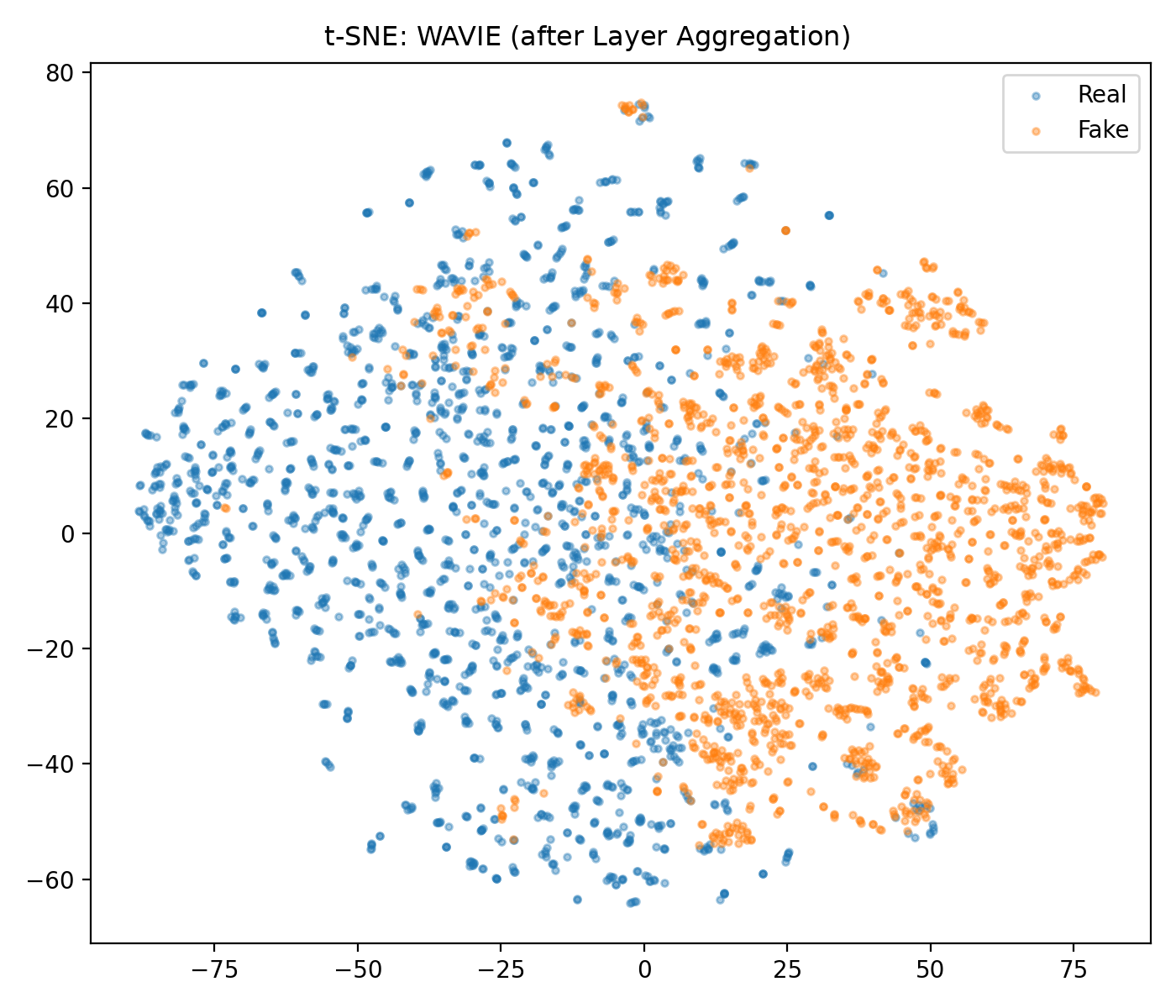}
    \captionsetup{font=footnotesize}
    \caption{v1: After Layer Aggregation}
    \label{fig:tsne_tie_cdfv1}
\end{subfigure}
\hfill
\begin{subfigure}[b]{0.45\linewidth}
    \includegraphics[width=\linewidth]{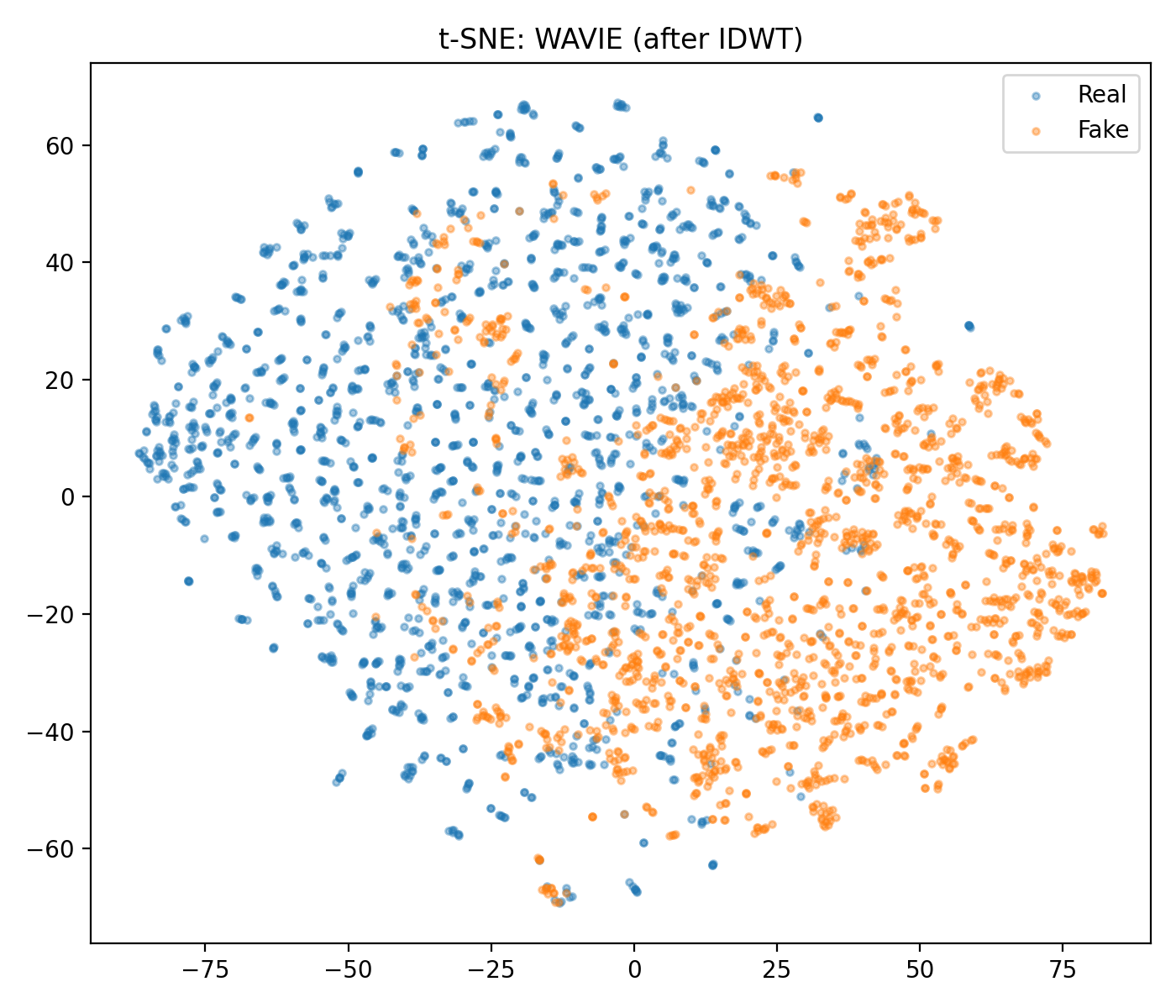}
    \captionsetup{font=footnotesize}
    \caption{v1: After IDWT}
    \label{fig:tsne_idwt_cdfv1}
\end{subfigure}

\vspace{4pt}

\begin{subfigure}[b]{0.45\linewidth}
    \includegraphics[width=\linewidth]{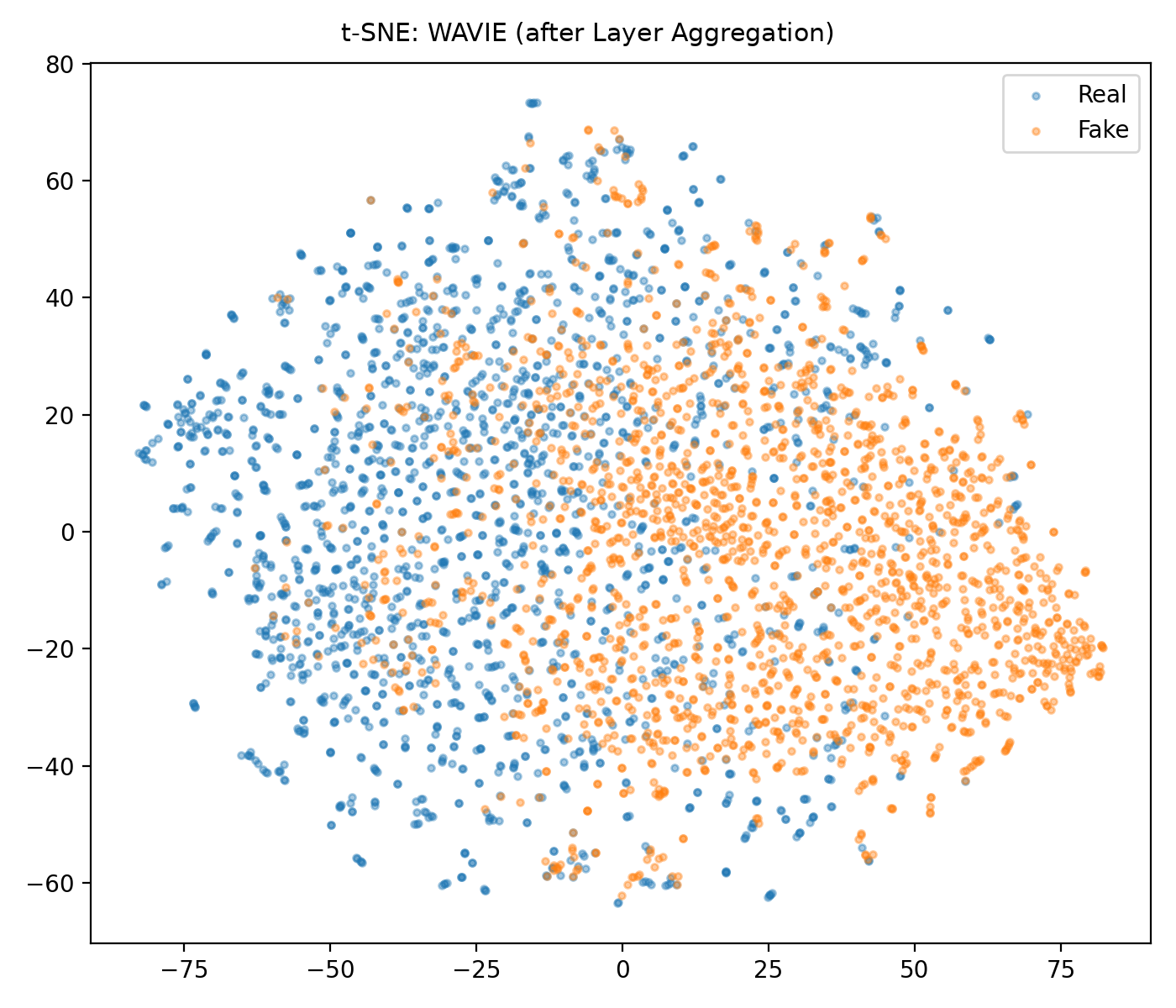}
    \captionsetup{font=footnotesize}
    \caption{v2: After Layer Aggregation}
    \label{fig:tsne_tie_cdfv2}
\end{subfigure}
\hfill
\begin{subfigure}[b]{0.45\linewidth}
    \includegraphics[width=\linewidth]{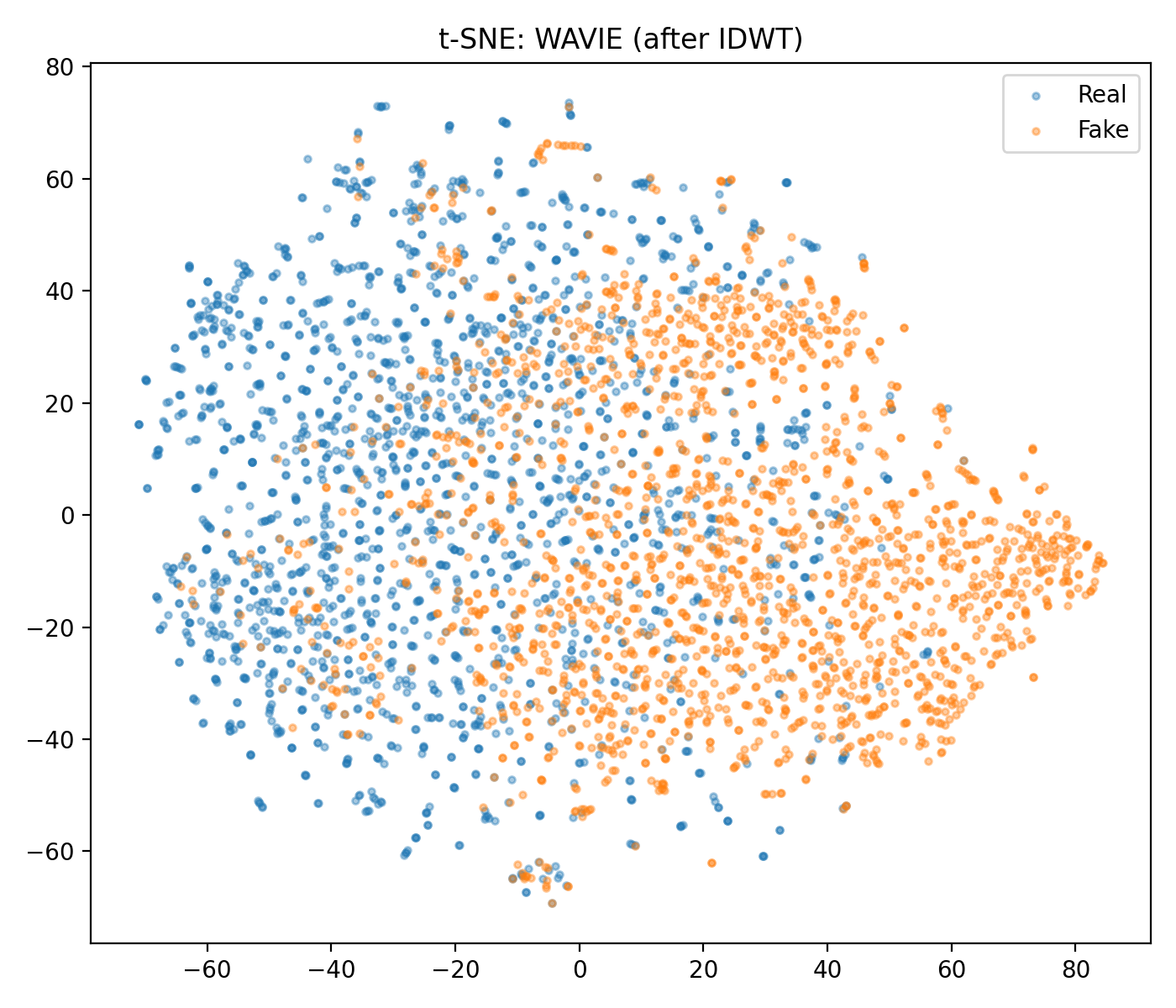}
    \captionsetup{font=footnotesize}
    \caption{v2: After IDWT}
    \label{fig:tsne_idwt_cdfv2}
\end{subfigure}
\captionsetup{font=footnotesize}
\caption{t-SNE visualizations after the Layer Aggregation (left) and IDWT (right) stages on Celeb-DF-v1 (top) and Celeb-DF-v2 (bottom). IDWT produces more separable and compact clusters.}
\label{fig:tsne_combined}
\end{figure}

\paragraph{t-SNE Evaluations.}
We visualize representation quality on \textbf{Celeb-DF-v1} and \textbf{Celeb-DF-v2} using t-SNE. For each split, we project both the intermediate (\textit{Layer Aggregated}) and final (\textit{IDWT}) embeddings produced by \textbf{WAVIE} into 2D. On Celeb-DF-v1, applying the complete IDWT yields a clearer gap between real and fake clusters; on Celeb-DF-v2, IDWT produces increased within-class density and tighter clusters relative to the Layer Aggregation stage. These trends, shown in Fig.~\ref{fig:tsne_combined}, indicate progressively more discriminative and compact representations as the pipeline advances. We use these plots as qualitative diagnostics; quantitative claims rely on AUROC.

\subsection{Ablation Studies}
To better understand the contribution of each component of WAVIE, we conduct ablation experiments on the Celeb-DF-v1 and v2 datasets. 
We evaluate three simplified variants:

\begin{itemize}
    \item \textbf{Without fusing intermediate features:} Instead of fusing intermediate encoder-block features, we directly pass the final CLIP ViT embedding into the wavelet module.
    \item \textbf{Without Wavelet Refinement:} We bypass the discrete wavelet transform and feed the aggregated feature (after $Q_{1}$) directly to the classifier.
    \item \textbf{Average Layer Aggregation:} Instead of aggregating CLS tokens using learnable weights, we take the average of the vectors fetched from the encoder blocks. 
\end{itemize}

\begin{table}[!t]
\centering
\begin{tabular}{lcc}
\toprule
Variant & Celeb-DF-v1 & Celeb-DF-v2 \\
\midrule
Without Fusion & 0.7532 & 0.7854 \\
Without Wavelet Refinement & 0.7221 & 0.7636 \\
Average Layer Aggregation & 0.8508 & 0.8404 \\
\textbf{WAVIE (Full)} & \textbf{0.8518} & \textbf{0.8520} \\
\bottomrule
\end{tabular}
\captionsetup{font=footnotesize}
\caption{Ablation Results (frame-level AUROC).}
\label{tab:ablations}
\end{table}

The results in Table~\ref{tab:ablations} highlight the role of each component. Removing the intermediate-layer fusion block (\emph{Without Fusion}) drops cross-domain AUROC by nearly $10$ points, underscoring the importance of intermediate transformer features. Eliminating the wavelet refinement (\emph{Without Wavelet}) further reduces performance, confirming that frequency-domain filtering plays a key role in cross-domain robustness. Uniform averaging (\emph{Average Layer Aggregation}) is competitive but consistently trails WAVIE, suggesting that learned per-layer weights provide a modest yet reliable benefit once the wavelet refinement is in place.

\section{CONCLUSION}\label{sec:conclusion}
We presented WAVIE, a parameter-efficient deepfake detection framework that jointly leverages multi-layer transformer representations and frequency-domain refinement for improved cross-domain generalization. The model remains competitive on FF++ ($\approx$0.98 AUROC) while markedly reducing the performance drop on unseen datasets, highlighting its suitability for deployment where manipulation methods are unknown. We believe that by leveraging intermediate-layer embeddings, WAVIE preserves fine-grained forgery cues often lost in final-layer representations, and the wavelet-based refinement further enhances robustness by emphasizing frequency inconsistencies introduced during manipulation.

Ablation studies confirm that both components are essential: removing intermediate fusion or wavelet refinement each cause substantial drops in cross-domain AUROC, while uniform layer averaging trails the learned aggregation weights, suggesting a practical efficiency--accuracy trade-off and a promising target for future compact variants. The total trainable parameter count remains low ($\sim$2M) atop a frozen backbone.

\noindent\textbf{Scope and Limitations:}
Our evaluation is frame-level and face-centric, trained exclusively on FF++ (c0) and tested on Celeb-DF and WildDeepFake. We employ a fixed 3-level db6 DWT without learning the filters and do not use image-space augmentations, which isolates the effect of our design but may limit robustness. Exploring learnable wavelet bases, adaptive decompositions, and stronger augmentations are promising directions.

\noindent\textbf{Future directions:}
Key avenues include: (i) Temporal modeling to exploit cross-frame artifacts; (ii) Learnable wavelet banks and joint spatial-frequency attention; and (iii) Robustness to compression and post-processing.

\section*{ACKNOWLEDGMENT}
The work is partly supported through project grant by Directorate of Forensic Science, India.

\addtolength{\textheight}{-1cm}   

\bibliographystyle{IEEEtran}
\bibliography{references}

\end{document}